\documentclass{article} 

\usepackage{preprint}

\usepackage{subfigure}
\usepackage[utf8]{inputenc} 
\usepackage[T1]{fontenc}    
\usepackage{hyperref}       
\usepackage{url}            
\usepackage{booktabs}       
\usepackage{amsfonts}       
\usepackage{nicefrac}       
\usepackage{mdwlist}
\usepackage{microtype}      

\usepackage{ulem}
\usepackage{xcolor,fixltx2e}

\newif\ifcomment\commenttrue
\ifcomment
    \newcommand{\pinaforecomment}[3]{\colorbox{#1}{\parbox{.8\linewidth}{#2: #3}}}
\else
    \newcommand{\pinaforecomment}[3]{}
\fi

\makeatletter
\newcommand{\setword}[2]{%
  \phantomsection
  #1\def\@currentlabel{\unexpanded{#1}}\label{#2}%
}
\makeatother
\usepackage{subfigure}
\usepackage{graphicx}
\newcommand*\rot{\rotatebox{90}}

\title{\textsc{ClimaText}: A Dataset for Climate Change Topic Detection}

\author{%
 Francesco Saverio Varini  \\
 Department of Computer Science\\
 ETH Zurich \\
  Zurich, Switzerland \\
 \texttt{fvarini@student.ethz.ch}
  \And
Jordan Boyd-Graber \\ 
CS, iSchool, LSC, and UMIACS\\
University of Maryland \\
College Park, MD, USA\\
\texttt{jbg@umiacs.umd.edu} 
  \And
  Massimiliano Ciaramita \\
  Google Research\\
  Zurich, Switzerland \\
  \texttt{massi@google.com} \\
 \And
 Markus Leippold \\
 Department of Banking and Finance \\
 University of Zurich \\
 Zurich, Switzerland \\
  \texttt{markus.leippold@bf.uzh.ch} 
}
\begin{document}

\maketitle

\begin{abstract}
Climate change communication in the mass media and other textual sources may affect and shape public perception. Extracting climate change information from these sources is an important task, e.g., for filtering content and e-discovery, sentiment analysis, automatic summarization, question-answering, and fact-checking. However, automating this process is a challenge, as climate change is a complex, fast-moving, and often ambiguous topic with scarce resources for popular text-based AI tasks. In this paper, we introduce \textsc{ClimaText}, a dataset for sentence-based climate change topic detection, which we make publicly available. We explore different approaches to identify the climate change topic in various text sources. We find that popular keyword-based models are not adequate for such a complex and evolving task. Context-based algorithms like BERT~\cite{devlin2018bert} can detect, in addition to many trivial cases, a variety of complex and implicit topic patterns. Nevertheless, our analysis reveals a great potential for improvement in several directions, such as, e.g., capturing the discussion on indirect effects of climate change. Hence, we hope this work can serve as a good starting point for further research on this topic.
\end{abstract}

\section{Introduction}
\label{sec:introduction}
The World Economic Forum~\cite{world_economic_report_2020} continues to rank climate change as one of the top global risks in the next ten years. Not surprisingly, climate change receives prominent public attention and media coverage, which makes it a fascinating object of study for natural language understanding (NLU). The first step in tasks such as sentiment analysis, fact-checking, and question-answering is the identification of the climate-change topic in text sources. This seems like an obvious task, which is commonly addressed by simple string matching from a keyword list; e.g., as in prominent financial economics literature~\cite{HedgingClimate}.
However, consider the following statements: 
\begin{enumerate*}
	\item Compliance with these laws and regulations could require significant commitments of capital toward environmental monitoring, renovation of storage facilities or transport vessels, payment of emission fees and carbon or other taxes, and application for, and holding of, permits and licenses.
	\item Al Gore's book is quite accurate, and far more accurate than contrarian books.
	\item The temperature is not rising nearly as fast as the alarmist computer models predicted.
	\item The parties also began discussing the post-Kyoto mechanism, on how to allocate emission reduction obligation following 2012, when the first commitment period ends.
	\item The rate of Antarctica ice mass loss has tripled in the last decade.
	\item Globally about $1\%$ of coral is dying out each year.
	\item Our landfill operations emit methane, which is identified as a GHG.\label{enum:10k_banal}
	\item Due to concerns about the de-forestation of tropical rain forests and climate change, many countries that have been the source of these hardwoods have implemented severe restrictions on the cutting and export of these woods.. \label{enum:10k_banal2}
	\item The 2015 conference was held at Le Bourget from 30 November to 12 December 2015.\label{enum:wiki_implicit}
	\item Polar bear numbers are increasing.\label{enum:sent_claims_amb}
	\item In 2006, CEI arranged a public-service television commercial about carbon dioxide with the slogan “They call it pollution; we call it life”\label{enum:sent_wiki_amb}
	\item Human activities emit about 29 billion tons of carbon dioxide per year, while volcanoes emit between 0.2 and 0.3 billion tons.\label{enum:sent_wiki_amb-2}
	\item Further, these emission control regulations could result in increased capital and operating costs.\label{enum:sent_10k_amb}
\end{enumerate*}

%
The first six sentences enumerated above cannot be detected using glossaries (Table~\ref{table:glossaries-keywords}) for keyword based models. These sentences are, instead, correctly identified using a machine-learned text classifier, BERT~\cite{devlin2018bert}. However, sentence \ref{enum:10k_banal} and \ref{enum:10k_banal2} are straightforward examples. Very rarely, for some reason, these kinds of sentences are not correctly classified by BERT according to our results. On the other hand, they are always detected correctly by the simplest keyword based models.
Climate change topic detection can be difficult and often patterns are implicit and ambiguous. For instance, while it is  implicit but unequivocal that sentence \ref{enum:wiki_implicit} talks about a climate change conference, it is unclear whether the sentences \ref{enum:sent_claims_amb} and \ref{enum:sent_wiki_amb} are about climate denialism, therefore about climate change. Sentence \ref{enum:sent_wiki_amb-2} talks only about a scientific fact: carbon dioxide emissions are there with or without climate change. Finally, sentence \ref{enum:sent_10k_amb} is talking about regulations on emissions, however, it is omitted which specific ones: emissions of oxygen, for instance, would not relate to climate change.

Climate change is a complex topic with many different facets. These facets can be textually described in different ways, potentially combining several words distributed within the text. Moreover, climate change is a fast-moving topic for which new terms and concepts are emerging, e.g., in the public debate and new legislation. Therefore, we need to catch up with the language used for it continually. We argue that NLU and machine learning are needed to solve this task, and that solutions will have broader societal value, by helping in keeping track of the topic and its ramifications.
To encourage research at the intersection of climate change and natural language understanding we built and make public a dataset for climate-related text classification together with preliminary findings.\footnote{Our data is made available on \url{www.climatefever.ai}. }


\section{Constructing the data set}\label{data}
The data consists of labeled sentences. The label indicates whether a sentence talks about climate change or not. Labels are generated heuristically or via a manual process. 
The manual labeling rules emerged through inspection of sentences and a collaborative labeling process with four raters, for which we monitored the inter-rater reliability through the Kappa statistic~\cite{kappa_coef} (Table~\ref{table:kappa-values}). We list the rules in Appendix \ref{appendix:sec:enumeration-labeling-rules}. 
Sentences are collected from different sources: Wikipedia, the U.S. Securities and Exchange Commission (SEC) 10K-files~\cite{SEC_filings}, which are annual regulatory filings in which listed companies in the US are required to self-identify climate-related risks that are material to their business, and a selection of climate-change claims collected from the web~\cite{claims_validation}.

For Wikipedia, we select documents through graph-based heuristics based on Wikipedia Inlinks (see Appendix \ref{appendix:wikilinks_graph}). We collect 6,885 documents, 715 relevant to climate change and 6,170 not relevant to climate change. We divide the documents between train, development, and test sets. Then, we split the documents into sentences, and we label these as climate change related or not, heuristically, using the same label as that of the document of origin (see Table~\ref{table:wikipedia-data-splits}).

\begin{table}[hpt]
  \caption{Wikipedia document labeled data sets (positives vs negatives in parentheses).}\label{table:wikipedia-data-splits}
  \centering
  \begin{tabular}{lll}
    \toprule
    \cmidrule(r){1-2}
    Data & Tag & Sentences \\
    \midrule
     Train split & \setword{\textbf{Wiki-Doc-Train}}{Data:Wiki-doc} & $115 854$ ($57 927$ vs $57 927$)     \\
     Development split & \setword{\textbf{Wiki-Doc-Dev}}{Data:Wiki-doc-dev} & $3 826$ ($1 913$ vs $1 913$)  \\
     Test split & \setword{\textbf{Wiki-Doc-Test}}{Data:Wiki-doc-test} & $3 826$ ($1 913$ vs $1 913$)  \\
    \bottomrule
  \end{tabular}
\end{table}

Training on the data from Table~\ref{table:wikipedia-data-splits} does not yield good predictive models 
because of the assumption that all sentences in a positive document are positives. Therefore, we follow up with Active Learning (AL)~\cite{AL_review} to manually label thousand of additional instances. For this purpose, we use DUALIST~\cite{Dualist}\cite{Dualist_followup}, a web-based framework performing AL and running a multinomial NB model in the loop (see Appendix \ref{appendix:AL_on_wiki_10ks}). We label sentences from \ref{Data:Wiki-doc}. We also run this labeling process on Item 1A of the 10-K files from 2014, as this is the relevant section in which climate risk must be reported. Table~\ref{table:AL-labeled-data} on the right side provides an overview of the data set created with AL.

\begin{table}[htp]        \caption{Evaluation and AL train sentences (positives vs negatives in parentheses)}
        \begin{minipage}[t]{0.5\textwidth}
\label{table:final-test-data} 
            \centering
            \begin{tabular}{cc}
                    \multicolumn{2}{c}{Evaluation sentences} \\
    \toprule
    \cmidrule(r){1-2}
 Data & Sentences \\ [0.5ex] 
 \midrule
 Wikipedia (dev) & $300$ ($79$ vs $221$)
 \\
 Wikipedia (test) & $300$ ($33$ vs $267$) \\ 
 10-Ks (2018, test) & $300$ ($67$ vs $233$)\\
 Claims (test) & $1000$ ($500$ vs $500$) \\
\bottomrule
\end{tabular}
\end{minipage}
    \begin{minipage}[t]{0.5\textwidth}
    \label{table:AL-labeled-data}
    \centering
        \begin{tabular}{lll}
        \multicolumn{3}{c}{Active learning train sentences} \\
        \toprule
    \cmidrule(r){1-2}
 Data & Tag & Sentences \\ [0.5ex]
    \midrule
 Wikipedia & \setword{\textbf{AL-Wiki}}{Data:AL-Wiki} & $3000$ ($261$ vs $2739$ ) \\
 10-Ks & \setword{\textbf{AL-10Ks}}{Data:AL-10Ks} & $3000$ ($58$ vs $2942$) \\
    \bottomrule
            \end{tabular}
        \end{minipage}
    \end{table}

For the evaluation data, we proceed as follows. First, we create a development set from \ref{Data:Wiki-doc-dev} and a test set from \ref{Data:Wiki-doc-test}. We sample 150 sentences from the positives and 150 from the negatives. The four raters then label these sentences according to the labeling rules. Each sentence is deemed negative only if all raters labeled it as negative, positive otherwise. Then, we create another test set using only the 10-K files by adopting a Wikipedia trained BERT-predictions-based sampling scheme, randomly selecting 150 examples both within the positive and negative predictions. Then, the four raters label these sentences according to the labeling rules. Lastly, we collect 500 positive and 500 negative claims from the sources used in~\cite{claims_validation}. The left side of Table~\ref{table:final-test-data} gives an overview of the development and test sets created.

\section{Analysis}
\label{models}
For our analysis of the dataset, we rely on three model frameworks for classification.

\paragraph{Keyword-based models:} We use several existing climate-related keywords sets as a benchmark, see Table \ref{table:glossaries-keywords}.

\begin{table}[h]
   \caption{Glossaries used for the keyword-based models}
    \label{table:glossaries-keywords}
  \centering
  \begin{tabular}{llr}
    \toprule
    \cmidrule(r){1-2}
 Keywords source & Tag & Number of Keywords\\ [0.5ex] 
    \midrule
 Wikipedia Glossary~\cite{wiki-glossary-cc} & \setword{\textbf{Wikipedia-Keywords}}{Model:Wiki-Keywords} & 175 \\
 IPCC Glossary~\cite{ipcc-glossary-cc} &  \setword{\textbf{IPCC-Keywords}}{Model:IPCC-Keywords} & 340\\
 Global Change Glossary~\cite{globalchange-glossary-cc} & \setword{\textbf{GlobalChange-Keywords}}{Model:GlobalChange-Keywords} & 126 \\
 FS-US Glossary~\cite{fedus-glossary-cc} & \setword{\textbf{FS-US-Keywords}}{Model:FSUS-Keywords} & 241 \\
 Small & \setword{\textbf{Small-Keywords}}{Model:Small-Keywords} & 6 \\
 All & \setword{\textbf{All-Keywords}}{Model:All-Keywords} & 771\\
    \bottomrule
  \end{tabular}
\end{table}

\paragraph{N{\"a}ive Bayes:} The N{\"a}ive Bayes (NB) classifier from DUALIST~\cite{Dualist}. NB models usually provide competitive baselines, though NB assumes independence of the features given the class.

\paragraph{BERT:} A popular attention-based text-classifier~\cite{devlin2018bert}. 
We use the BERT\textsubscript{BASE} model pre-trained on Wikipedia and fine-tune it by adding an output layer for our specific binary classification task.

\section{Results and discussion}
\label{headings}

\begin{figure}[htbp]  \label{fig:all_test_results}
\centering  
\resizebox{0.95\columnwidth}{!}{%
\includegraphics[width=0.85\textwidth]{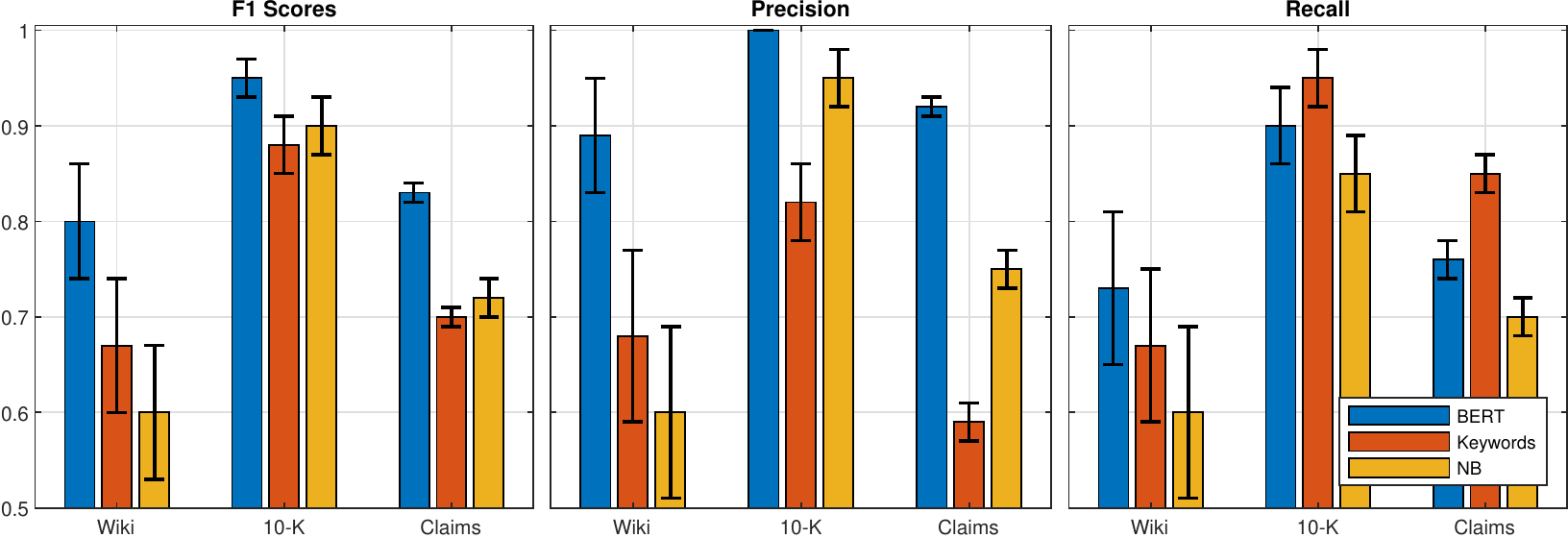}%
}
\caption{ \label{fig:all_results}}
The figure reports the highest F1-score (left panel) with the respective precision (middle panel) and recall (right panel) for the BERT, the keyword-based, and NB models.
\end{figure}

We present results on the test sets, using mean accuracy, precision, recall, and F1 as performance metrics, estimating the standard error via bootstrapping.\footnote{BERT and NB are optimized with respect to F1 on the Wikipedia development set in Table~\ref{table:final-test-data}.} Our results are as follows. First, as Figure~\ref{fig:all_results} suggests, in terms of the F1 score, BERT outperforms the other models for all three test sets. The NB classifier, in contrast, fails to beat the keywords-based approach for the Wikipedia test set. Second, the best BERT model achieves the highest precision for all test sets, see Figure~\ref{fig:all_results}. Indeed, BERT's performance in precision is the main cause for the performance in terms of F1.\footnote{Tables~\ref{table:claims-test-results} to \ref{table:wikipedia-test-results} in Appendix \ref{appendix:extensive_results} provide a detailed overview on the test results.} Third, while the BERT models reach a precision around or above $90\%$  (see Figure~\ref{fig:all_results}), and maintain a remarkable recall higher or equal to $73\%$, the keyword approach has better recall on two of the three tasks.

We also find that the 10-K files are well standardized and present climate change topic patterns that are easier to detect than in Wikipedia or the collected claims. Indeed, these regulatory reports are often criticized for being boilerplate.\footnote{However, using BERT, \cite{bertLeippold} find that 10-K filings do include important climate risk information that influence prices in financial markets.} However, we should remark that the 10K test set is sampled conditioning on the predictions of a BERT model. This means that it might be not entirely representative of the climate change topic distribution in the 10K files.

The above findings indicate that the climate change topic detection task in sentences can be challenging, even for state-of-the-art neural network models such as BERT. Keyword-based models are outperformed by BERT in terms of precision, but can be competitive in terms of recall. 
Can keyword-based models, being deterministic models, rise to the challenge of climate change language/patterns shifts over time? In principle, we could periodically enlarge the keywords set through automatic keywords discovery methods~\cite{Beliga20141KE}\cite{keyword_review}. However, such a procedure may not work when the topic is as complex as climate change. We may need to discover combinations of words in the same sentence rather than single keywords/key-phrases. Therefore, one may need an oracle intervention, like a human encoding complicated patterns periodically in the deterministic search algorithm. 



In the future, we plan to enlarge the data sets to eventually include a wider variety of the representative climate change topic distribution in sentences. 
An exciting avenue of further research is to 
understand the challenges for models like BERT when applied to complex topics like climate change. Finally, while we currently use sentences as the unit of analysis, a contextual understanding of whether a text is climate-related would be more nuanced and incorporate contextual information.

\section*{Acknowledgments}

Research supported with Cloud TPUs from Google's TensorFlow Research Cloud (TFRC).

\bibliographystyle{IEEEtran}
\bibliography{refs}

\newpage

\appendix

\section{Test Results}\label{appendix:extensive_results}

 In the following results tables the model name is flagged with the training data used. In particular, whenever an ``\&'' is present in a model name, it means that we train the model on successive training steps using different data. From left to right, the model training data is listed in the exact order of usage.
 
\begin{table}[h]
    \caption{Claims bootstrap test results sorted by F1 score descending. Standard deviation in parenthesis.}\label{table:claims-test-results}
  \centering
  \resizebox{\columnwidth}{!}{%
  \begin{tabular}{lllll}
    \toprule
    \cmidrule(r){1-2}
Model & \rot{Accuracy} & \rot{F1} & \rot{Precision} & \rot{Recall}\\ [0.5ex] 
    \midrule
BERT-\ref{Data:Wiki-doc} \& \ref{Data:AL-10Ks} $\cup$ \ref{Data:AL-Wiki} & 0.85 (0.01) & 0.83 (0.01) & 0.92 (0.01) & 0.76 (0.02) \\
BERT-\ref{Data:Wiki-doc} \& \ref{Data:AL-10Ks} \& \ref{Data:AL-Wiki} & 0.84 (0.01) & 0.82 (0.01) & 0.95 (0.01) & 0.73 (0.02) \\
BERT-\ref{Data:AL-Wiki} & 0.81 (0.01) & 0.81 (0.01) & 0.83 (0.02) & 0.79 (0.02) \\
BERT-\ref{Data:AL-10Ks} \& \ref{Data:AL-Wiki} & 0.82 (0.01) & 0.81 (0.01) & 0.89 (0.01) & 0.74 (0.02) \\
BERT-\ref{Data:Wiki-doc} \& \ref{Data:AL-10Ks} & 0.82 (0.01) & 0.8 (0.01) & 0.92 (0.01) & 0.72 (0.02) \\
BERT-\ref{Data:Wiki-doc} \& \ref{Data:AL-Wiki} & 0.83 (0.01) & 0.8 (0.01) & 0.96 (0.01) & 0.69 (0.02) \\
BERT-\ref{Data:Wiki-doc} \& \ref{Data:AL-Wiki} \& \ref{Data:AL-10Ks} & 0.82 (0.01) & 0.79 (0.01) & 0.92 (0.01) & 0.7 (0.02) \\
NB-\ref{Data:AL-Wiki} & 0.73 (0.01) & 0.72 (0.02) & 0.75 (0.02) & 0.7 (0.02) \\
BERT-\ref{Data:Wiki-doc} & 0.62 (0.02) & 0.72 (0.01) & 0.57 (0.02) & 0.98 (0.01) \\
BERT-\ref{Data:AL-Wiki} \& \ref{Data:AL-10Ks} & 0.76 (0.01) & 0.72 (0.02) & 0.89 (0.02) & 0.6 (0.02) \\
\ref{Model:All-Keywords} & 0.63 (0.01) & 0.7 (0.01) & 0.59 (0.02) & 0.85 (0.02) \\
\ref{Model:IPCC-Keywords} & 0.62 (0.02) & 0.67 (0.02) & 0.59 (0.02) & 0.78 (0.02) \\
\ref{Model:Wiki-Keywords} & 0.72 (0.01) & 0.64 (0.02) & 0.9 (0.02) & 0.5 (0.02) \\
BERT-\ref{Data:AL-10Ks} & 0.71 (0.01) & 0.64 (0.02) & 0.85 (0.02) & 0.51 (0.02) \\
\ref{Model:FSUS-Keywords} & 0.69 (0.01) & 0.62 (0.02) & 0.81 (0.02) & 0.5 (0.02) \\
NB-\ref{Data:AL-10Ks} & 0.69 (0.01) & 0.6 (0.02) & 0.82 (0.02) & 0.47 (0.02) \\
\ref{Model:GlobalChange-Keywords} & 0.62 (0.02) & 0.41 (0.02) & 0.88 (0.03) & 0.27 (0.02) \\
\ref{Model:Small-Keywords} & 0.6 (0.02) & 0.34 (0.02) & 1.0 (0.0) & 0.2 (0.02) \\
    \bottomrule
  \end{tabular}%
  }
\end{table}

\begin{table}
    \caption{10-K bootstrap test results sorted by F1 score descending. Standard deviation in parenthesis.}
    \label{table:10K-test-results}
  \centering
\resizebox{\columnwidth}{!}{%
  \begin{tabular}{lllll}
    \toprule
    \cmidrule(r){1-2}
Model & \rot{Accuracy} & \rot{F1} & \rot{Precision} & \rot{Recall}\\ [0.5ex] 
    \midrule
BERT-\ref{Data:Wiki-doc} \& \ref{Data:AL-10Ks} \& \ref{Data:AL-Wiki} & 0.98 (0.01) & 0.95 (0.02) & 1.0 (0.0) & 0.9 (0.04) \\
BERT-\ref{Data:AL-Wiki} \& \ref{Data:AL-10Ks} & 0.97 (0.01) & 0.93 (0.02) & 0.92 (0.03) & 0.93 (0.03) \\
BERT-\ref{Data:Wiki-doc} \& \ref{Data:AL-10Ks} $\cup$ \ref{Data:AL-Wiki} & 0.97 (0.01) & 0.92 (0.02) & 0.95 (0.03) & 0.9 (0.04) \\
BERT-\ref{Data:Wiki-doc} \& \ref{Data:AL-Wiki} \& \ref{Data:AL-10Ks} & 0.97 (0.01) & 0.92 (0.02) & 0.95 (0.03) & 0.9 (0.04) \\
BERT-\ref{Data:Wiki-doc} \& \ref{Data:AL-10Ks} & 0.97 (0.01) & 0.92 (0.02) & 0.95 (0.03) & 0.89 (0.04) \\
BERT-\ref{Data:AL-10Ks} \& \ref{Data:AL-Wiki} & 0.96 (0.01) & 0.92 (0.03) & 0.91 (0.03) & 0.93 (0.03) \\
NB-\ref{Data:AL-10Ks} & 0.96 (0.01) & 0.9 (0.03) & 0.95 (0.03) & 0.85 (0.04) \\
\ref{Model:FSUS-Keywords} & 0.94 (0.01) & 0.88 (0.03) & 0.82 (0.04) & 0.95 (0.03) \\
BERT-\ref{Data:Wiki-doc} \& \ref{Data:AL-Wiki} & 0.95 (0.01) & 0.87 (0.03) & 0.98 (0.02) & 0.79 (0.05) \\
BERT-\ref{Data:AL-10Ks} & 0.94 (0.01) & 0.86 (0.03) & 0.82 (0.04) & 0.91 (0.03) \\
\ref{Model:Small-Keywords} & 0.94 (0.01) & 0.85 (0.04) & 1.0 (0.0) & 0.74 (0.05) \\
\ref{Model:Wiki-Keywords} & 0.91 (0.02) & 0.79 (0.04) & 0.83 (0.05) & 0.75 (0.05) \\
BERT-\ref{Data:AL-Wiki} & 0.83 (0.02) & 0.71 (0.04) & 0.58 (0.05) & 0.94 (0.03) \\
\ref{Model:GlobalChange-Keywords} & 0.8 (0.02) & 0.56 (0.05) & 0.56 (0.06) & 0.57 (0.06) \\
NB-\ref{Data:AL-Wiki} & 0.59 (0.03) & 0.5 (0.04) & 0.34 (0.04) & 0.91 (0.04) \\
\ref{Model:All-Keywords} & 0.44 (0.03) & 0.44 (0.04) & 0.28 (0.03) & 1.0 (0.0) \\
BERT-\ref{Data:Wiki-doc} & 0.39 (0.03) & 0.42 (0.04) & 0.27 (0.03) & 1.0 (0.0) \\
\ref{Model:IPCC-Keywords} & 0.44 (0.03) & 0.42 (0.04) & 0.27 (0.03) & 0.91 (0.04) \\
    \bottomrule
  \end{tabular}%
  }
\end{table}

\begin{table}
    \caption{Wikipedia bootstrap test results sorted by F1 score descending. Standard deviation in parenthesis.}
    \label{table:wikipedia-test-results}
  \centering
\resizebox{\columnwidth}{!}{%
  \begin{tabular}{lllll}
    \toprule
    \cmidrule(r){1-2}
Model & \rot{Accuracy} & \rot{F1} & \rot{Precision} & \rot{Recall}\\ [0.5ex] 
    \midrule
BERT-\ref{Data:AL-Wiki} \& \ref{Data:AL-10Ks} & 0.96 (0.01) & 0.8 (0.06) & 0.89 (0.06) & 0.73 (0.08) \\
BERT-\ref{Data:Wiki-doc} \& \ref{Data:AL-Wiki} & 0.95 (0.01) & 0.79 (0.06) & 0.79 (0.07) & 0.79 (0.07) \\
BERT-\ref{Data:AL-10Ks} \& \ref{Data:AL-Wiki} & 0.95 (0.01) & 0.77 (0.06) & 0.76 (0.07) & 0.79 (0.07) \\
BERT-\ref{Data:Wiki-doc} \& \ref{Data:AL-10Ks} $\cup$ \ref{Data:AL-Wiki} & 0.94 (0.01) & 0.75 (0.06) & 0.69 (0.08) & 0.82 (0.07) \\
BERT-\ref{Data:Wiki-doc} \& \ref{Data:AL-Wiki} \& \ref{Data:AL-10Ks} & 0.94 (0.01) & 0.75 (0.06) & 0.74 (0.08) & 0.76 (0.08) \\
BERT-\ref{Data:Wiki-doc} \& \ref{Data:AL-10Ks} \& \ref{Data:AL-Wiki} & 0.93 (0.01) & 0.71 (0.06) & 0.67 (0.08) & 0.76 (0.08) \\
BERT-\ref{Data:AL-Wiki} & 0.92 (0.02) & 0.69 (0.06) & 0.6 (0.08) & 0.82 (0.07) \\
BERT-\ref{Data:AL-10Ks} & 0.92 (0.02) & 0.68 (0.06) & 0.63 (0.08) & 0.73 (0.08) \\
\ref{Model:Wiki-Keywords} & 0.93 (0.01) & 0.67 (0.07) & 0.68 (0.09) & 0.67 (0.08) \\
BERT-\ref{Data:Wiki-doc} \& \ref{Data:AL-10Ks} & 0.91 (0.02) & 0.66 (0.06) & 0.58 (0.08) & 0.79 (0.07) \\
NB-\ref{Data:AL-10Ks} & 0.91 (0.02) & 0.6 (0.07) & 0.6 (0.09) & 0.6 (0.09) \\
\ref{Model:FSUS-Keywords} & 0.88 (0.02) & 0.57 (0.07) & 0.47 (0.07) & 0.73 (0.08) \\
NB-\ref{Data:AL-Wiki} & 0.84 (0.02) & 0.55 (0.06) & 0.4 (0.06) & 0.88 (0.06) \\
\ref{Model:Small-Keywords} & 0.92 (0.02) & 0.46 (0.1) & 1.0 (0.0) & 0.3 (0.08) \\
BERT-\ref{Data:Wiki-doc} & 0.63 (0.03) & 0.37 (0.05) & 0.23 (0.04) & 1.0 (0.0) \\
\ref{Model:GlobalChange-Keywords} & 0.88 (0.02) & 0.34 (0.08) & 0.41 (0.11) & 0.3 (0.08) \\
\ref{Model:All-Keywords} & 0.52 (0.03) & 0.29 (0.04) & 0.17 (0.03) & 0.91 (0.05) \\
\ref{Model:IPCC-Keywords} & 0.55 (0.03) & 0.28 (0.04) & 0.17 (0.03) & 0.81 (0.07) \\
    \bottomrule
  \end{tabular}%
  }
\end{table}

\newpage

\section{Final Labeling Rules}

The labeling rules we agreed upon are the following:

\begin{enumerate}\label{appendix:sec:enumeration-labeling-rules}
    \item The sentence labeled as positive must talk about climate change.
    \begin{enumerate}
         \item Just discussing nature /  environment is not sufficient. 
         \item Discussing a general scientific fact or describing an aspect of the climate is only relevant if it is a mechanism / cause / effect of (climate) change.
         \begin{enumerate}
             \item No: “Methane is CH4”
             \item No: “Monsoons can affect  shipping”
             \item Yes: “Methane increases   temperature”
             \item Yes:  “The Monsoon season could be more volatile than the last century”
         \end{enumerate}
        \item “Change” must be an aggregate change over longer periods of time
        \item Just mentioning clean energy, emissions, fossil fuels, etc. is not sufficient
        \begin{enumerate}
        \item rather it must be connected to an environmental (CO2) 
        \item or societal aspect (divestment,  Kyoto  treaty) of climate change.
        \end{enumerate}
        \item Acid rain / pollution / etc. are environmental issues but are not related to climate change.
        \item Acronyms or names of entities, potentially well connected to climate change, must be mentioned along with some mechanism/cause/effect of climate change
        \begin{enumerate}
            \item No: “EPA has adopted new regulations”
            \item Yes: “EPA has adopted regulations in response to findings on increased    emissions of carbon dioxide” 
        \end{enumerate}
        \end{enumerate}
    \item The sentence can talk about climate change during any period of Earth’s history.
    \begin{enumerate}
        \item Yes: Massive eruptions all over the Earth’s surface caused lower temperatures for the next few centuries.
    \end{enumerate}
    \item There may be ambiguity because we only consider individual sentences.
    \begin{enumerate}
        \item If you cannot resolve an ambiguous reference (is EPA European Pressphoto Agency or Environmental Protection Agency), then use your best judgement about how to resolve the reference.
        \item If you don’t know what a person, event, or idea is, you can expand your knowledge with a quick web search.
        \item If after a quick quick search you still do not understand or in all other cases, label it as not relevant.
    \end{enumerate}
    \item In case of doubt and in all the other cases, the sentence must be labeled as negative.
\end{enumerate}

\begin{table}[h]
    \caption{Kappa coefficients and translated agreement level~\cite{kappa_coef}}\label{table:kappa-values}
  \centering
  \begin{tabular}{ll}
    \toprule
    \cmidrule(r){1-2}
     Kappa's value range & Agreement level \\    
     \midrule
     {[}.0, .20{]} & None 
     \\ 
     {[}.21, .39{]} & Minimal
     \\
     {[}.49, .59{]} & Weak 
     \\
     {[}.60, .79{]} & Moderate 
     \\
     {[}.80, .90{]} & Strong 
     \\ 
     {[}.90, 1{]} & (Almost) perfect 
     \\ 
    \bottomrule
  \end{tabular}
\end{table}

\section{Wikipedia inlinks graph}\label{appendix:wikilinks_graph}

For Wikipedia, we perform a document selection through graph-based heuristics, which we describe below. The selection procedure builds on the Normalized Google Distance~\cite{cilibrasi2004google}, which is given by:

\begin{equation} \label{math:GSD}
sr(a,b)=\frac{\log(\max(|A|, |B|))-\log(|A\cap B|)}{\log(|W|)-\log(\min(|A|,|B|))}
\end{equation}

The idea behind the Normalized Google Distance in equation (\ref{math:GSD}) is to establish a similarity score between a pair of documents. The intuition is to base this similarity on how many other articles link both of the documents in the pair, as in \cite{Wikilinks_similarity}.
This is signaled in equation (\ref{math:GSD}) by the intersection in the numerator of the equation $\log(|A\cap B|)$. The rest of the mathematical terms are just part of the normalization. For normalization the NGD takes into account both the cardinality of the link sets to article $a$ and $b$ and the total size of Wikipedia. The Normalized Google Distance usually assigns a similarity score between 0 (identical) and 1 (unrelated). We can notice that the similarity score jumps to negative infinity when the articles \emph{a} and \emph{b} are linked by two distinct non-empty sets of articles (\emph{A} intersected with \emph{B} is empty). This is due to the numerator set intersection in the NGD formula.

Now that it is clear what the Normalized Google Distance is, we explain in detail how we apply it to our document selection problem. The procedure we follow consists of a lot of pre-computation to avoid being stuck with expensive calculation and, possibly, out of memory errors. Basically, the first step is to construct four dictionaries from the 01/11/2019 Wikipedia dumps~\cite{Wikidumps}: 
\begin{enumerate}
    \item A ``Title To Integer'' dictionary mapping each article title to an integer\label{item:MapToInteger}
    \item A ``Integer To Title'' dictionary mapping an integer to the respective article title\label{item:MapToTitle}
    \item A ``is Linking'' dictionary mapping a certain article title to the set of this article links. Each of the link is an article title\label{item:isLinking}
    \item A ``is Linked By'' dictionary mapping an article title to the title of the articles linking it.\label{item:isLinkedBy}
\end{enumerate}
The dictionaries in \ref{item:isLinking} and  \ref{item:isLinkedBy} contain the articles title mapped to integers by the dictionary in \ref{item:MapToInteger}. This in order to avoid out of memory problems when loading the dictionaries on a single machine.

\begin{figure}[ht]
\centering
\includegraphics[scale=1.1]{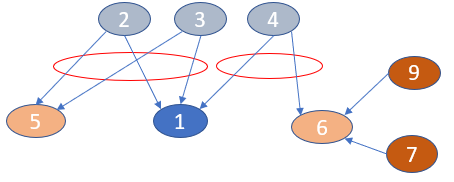}
\caption{\label{fig:globalwarming-link-graph}}
A toy illustration of the links graph in Wikipedia.
\end{figure}

At this point, using the dictionaries in \ref{item:isLinking} and \ref{item:isLinkedBy}, we are ready to traverse the Wikipedia articles graph. We are interested in the articles related to climate change. 
When we start our work, the Wikipedia  ``Climate change'' article is a redirection page to ``Global warming''. For this reason, though we know that climate change is a wider topic than global warming, we decide to start our traversal of the graph from ``Global warming''. From such a Wikipedia links graph, the goal is to find similar documents to ``Global warming''. The similarity score between pairs of documents is calculated with the NGD. In this regard, once again, we notice that if two articles are co-linked by no other article, then the NGD is negative infinity. This means that we do not really need to compute all possible pairs of articles similarity, but potentially only between co-linked pairs of articles. Conscious of this, we follow these steps for the graph traversal, which are represented in Figure \ref{fig:globalwarming-link-graph}:

\label{subsubsec:wikilinks_procedure}

\begin{enumerate}
    \item We pick a set of articles $S_{a}$ to start with, which is composed of only ``Global warming'' initially. In the Figure \ref{fig:globalwarming-link-graph} this is represented by the node 1 in blue.\label{item:wikilinks_step1}
    
    \item We collect the set of articles $A_{a}$ linking the article $a$ for each $a$ in  $S_{a}$ by using the dictionary in \ref{item:isLinkedBy}. We store the results in a dictionary \emph{$S_{a}$-to-$A_{a}$}. In the Figure \ref{fig:globalwarming-link-graph} the newly collected articles in this step are represented by the nodes in gray.\label{item:wikilinks_step2}
    
    \item For all the articles in each $A_{a}$ stored in \emph{$S_{a}$-to-$A_{a}$}, we collect the articles linked by them, excluding the ones in $S_{a}$, using the dictionary in \ref{item:isLinking}. These are the yellow nodes in Figure \ref{fig:globalwarming-link-graph}. We call the retrieved set of articles $S_{b}$ and we create a dictionary mapping each $b$ to all articles $a$ from $S_{a}$ which have at least a parent article in common. We call this dictionary \emph{$b$-to-$SS_{a}$}, where $SS_{a}$ stands for a subset of the set $S_{a}$. This last dictionary is useful for us during iterations other than the initial one, where $S_{a}$ is composed by more than one article. \label{item:wikilinks_step3}
    
    \item Next, we gather the set of articles $B_{b}$ linking article $b$ for each $b$ in $S_{b}$ by using the dictionary in \ref{item:isLinkedBy}. We store the results in a dictionary \emph{$S_{b}$-to-$B_{b}$}. In the Figure \ref{fig:globalwarming-link-graph} the newly collected articles in this step are represented by the nodes in brown and gray.\label{item:wikilinks_step4}

    \item Now, we calculate the Normalized Google Distance where, looking at the formula in \ref{math:GSD}, $a$ and $b$ comes respectively from the articles in $S_{a}$ and $S_{b}$ and $A$ and $B$ comes respectively from \emph{$S_{a}$-to-$A_{a}$} and \emph{$S_{b}$-to-$B_{b}$}. In our toy example from Figure \ref{fig:globalwarming-link-graph}, we will calculate the NGD between the blue article node in $S_{a}$ and each of the yellow article nodes in $S_{b}$. The process can then repeat from \ref{item:wikilinks_step2}, setting the new $S_{a}$ as the newly retrieved $S_{b}$, which become our new blue reference nodes.\label{item:wikilinks_step5}
\end{enumerate}\label{enum:wikilinks-steps}

We can notice  from Figure \ref{fig:globalwarming-link-graph} that the set of articles linking to article $5$, namely articles $2$ and $3$, is a subset of the articles linking article $1$. This is signaled by the red circle on the left of Figure \ref{fig:globalwarming-link-graph}. The same situation does not happen for article $6$, which has only article $4$ as a common parent with article $1$. Thus, we expect the NGD calculated between article 1 and 6 ($NGD(1,6)$) to be lower, therefore better, than the one between article 1 and 5 ($NGD(1,5)$). In fact, if we made the calculation in Formula \ref{math:GSD}, the $NGD(1,5)$ would have a lower numerator, due to a higher cardinality of the sets intersection. At the same time, we would find that $NGD(1,5)$ has a higher denominator than $NGD(1,6)$, since article $5$ is linked by two articles, which is lower than the three articles linking $6$. Therefore $\log(|W|)-\log(\min(|A|,|B|))$ is higher in $NGD(1,5)$ than in $NGD(1,6)$.

We need to make still few important remarks about the procedure followed:
\begin{enumerate}
    \item First, for each article collected during the graph traversal, we attribute a unique NGD similarity coefficient with respect to only one article. The problem is, in fact, that we can find the same article $b$ in $S_{b}$ which has common parent articles with more articles $a$ from $S_{a}$ in step \ref{item:wikilinks_step4} of the algorithm. In addition, we can still find the same article $b$ in $S_{b}$ in successive iterations. Therefore, we decide to select only the minimum score at the minimum distance level from the top ``Global warming'' article. We assume that the less distant linking-wise is an article to the starting article (``Global warming''), the more related to the latter it is.
    \item Second, as the computation complexity grows exponentially in the number of collected articles $N$ at each completed iteration, we decide to threshold the NGD similarity coefficients obtained. We decide to keep only the articles in $S_{b}$ whose pair similarity is below this threshold. We collect these articles in the new $S_{a}$ for the new iteration.
\end{enumerate}

 We run this algorithm to select the positive documents of the data set. Then, we sample the negative documents at random from the entire Wikipedia articles collection.

\section{Active Learning}
\label{appendix:AL_on_wiki_10ks}

The main motivation behind choosing an active learning~\cite{AL_review} algorithm is to try to achieve greater accuracy with fewer samples. This is accomplished by letting the model choose which instance/feature to label.

\begin{figure}[ht]
\centering
\includegraphics[width=0.8\columnwidth]{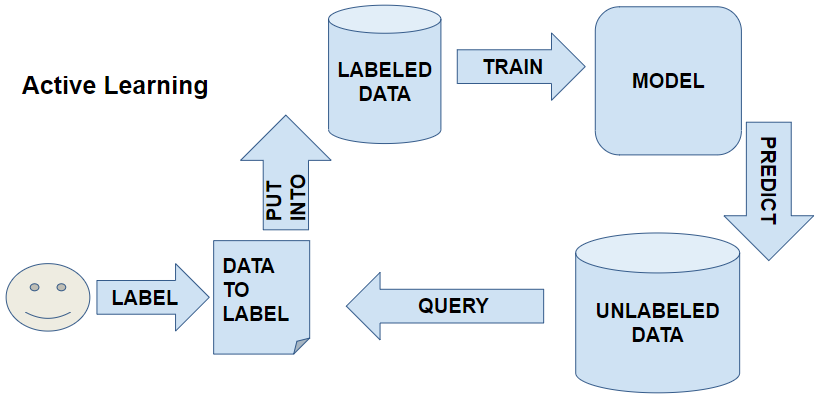}
\caption{\label{fig:al-procedure}}
Active Learning in one picture.
\end{figure}

The high level procedure of active learning is shown in Figure \ref{fig:al-procedure} and it is described as follows:
\begin{enumerate}
    \item Initialize a model or make use of an already trained model. Fit the model on a (new) set of labeled data, if any is currently available.\label{enum:al-steps-1}
    \item Predict with the model on the unlabeled data.
    \item Use the model predictions to pose queries on the unlabeled data.
    \item Show the data queried to an oracle (e.g. human) to be labeled
    \item Once the queried data is labeled, put it into the labeled data, while removing it from the unlabeled set.
    Repeat the process from \ref{enum:al-steps-1}.
\end{enumerate}\label{enum:al-steps}

\section{DUALIST active queries explained}\label{appendix:dualist-details}

\begin{equation}\label{equation:entropy-based-us}
H_\theta (Y\:|\:x) = -\sum_{j}^{}P_\theta (y_j\:|\:x)*\log P_\theta (y_j\:|\:x)
\end{equation}

\begin{figure}[ht]
\centering
\includegraphics[scale=.8]{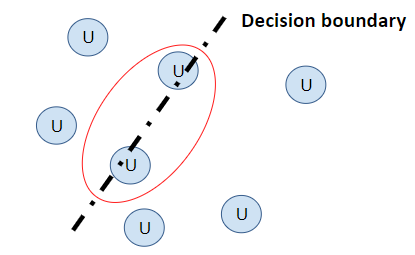}
\caption{\label{fig:model-decision-boundary}}
A two dimensional representation of the model decision boundary and the location of different samples. The red circle indicates the closest samples to the decision boundary
\end{figure}

DUALIST pose queries on instances and features. The features are unigram and bi-gram from the sentences.
Equation (\ref{equation:entropy-based-us}) represent the entropy based uncertainty sampling which DUALIST uses to query the instances to label by the oracle. Given the model predictions, we can attribute an entropy score to each unlabeled sentence from the unlabeled set. Then we can rank the sentences according to their entropy score. The higher the score, the more confused the model is about their classification. This usually happens to be around the model decision boundary as displayed in a toy two-dimensional space in Figure \ref{fig:model-decision-boundary}. However, there is no guarantee that these samples are the most informative for the task at hand. In fact, they could also be outliers, meaning samples which deviates significantly from the rest of them. Outliers confuse the model and make the training, together with the labeling, inefficient.

\begin{equation} \label{equation:ig}
IG(f\textsubscript{k}) = \sum_{I\textsubscript{k}}^{}\sum_{j}{}P(I\textsubscript{k},\:y\textsubscript{j})*\log\frac{P(I\textsubscript{k},\:y\textsubscript{j})}{P(I\textsubscript{k})*P(y\textsubscript{j})}
\end{equation}

As we previously mentioned, DUALIST queries also the features to label by the oracle.
Equation (\ref{equation:ig}) represents the information gain mathematical expression which DUALIST uses to score the features. To calculate the score, we need to know the label of our sentences. For this purpose, DUALIST makes use of the data labeled that far, if any, and the probabilistically-labeled instances by the model predictions on the unlabeled data. In equation (\ref{equation:ig}), it is taken the ratio between the joint probability of a feature \emph{k} to occur in a sample with label $j$, with the chance the feature and the label occur independently of each other. This is shown in the logarithmic part on the right hand side of the equation. We can also observe that the IG scores are per class $j$, which means that each class has its own feature importance ranking.

\end{document}